\pdfoutput=1

\documentclass[11pt]{article}

\usepackage[]{acl}

\usepackage{times}
\usepackage{latexsym}
\usepackage{listings}

\usepackage[T1]{fontenc}

\usepackage[utf8]{inputenc}

\usepackage{microtype}

\usepackage{amsfonts}
\usepackage{booktabs}
\usepackage{siunitx}
\usepackage{graphicx}
\usepackage{multicol}        
\usepackage{multirow}
\usepackage{amsmath}

\usepackage{listings}
\lstdefinestyle{mystyle}{
    basicstyle=\ttfamily\footnotesize,
}
\title{Retrieval-Augmented Neural Response Generation \\ Using Logical Reasoning and Relevance Scoring}

\author{Nicholas Thomas Walker \\
  Norwegian Computing Center \\
  Oslo, Norway \\
  \texttt{walker@nr.no} \\\And
  Stefan Ultes \\
  University of Bamberg \\
  Bamberg, Germany \\
  \texttt{stefan.ultes@uni-bamberg.de} \\
  \And
  Pierre Lison \\
  Norwegian Computing Center \\
  Oslo, Norway \\
  \texttt{plison@nr.no} \\}

\begin{document}
\maketitle
\begin{abstract}
Constructing responses in task-oriented dialogue systems typically relies on information sources such the current dialogue state or external databases. This paper presents a novel approach to knowledge-grounded response generation that combines retrieval-augmented language models with logical reasoning. The approach revolves around a knowledge graph representing the current dialogue state and background information, and proceeds in three steps. The knowledge graph is first enriched with logically derived facts inferred using probabilistic logical programming. A neural model is then employed at each turn to score the conversational relevance of each node and edge of this extended graph. Finally, the elements with highest relevance scores are converted to a natural language form, and are integrated into the prompt for the neural conversational model employed to generate the system response. 

We investigate the benefits of the proposed approach on two datasets (KVRET and GraphWOZ) along with a human evaluation. Experimental results show that the combination of (probabilistic) logical reasoning with conversational relevance scoring does increase both the factuality and fluency of the responses.
\end{abstract}

\section{Introduction}


Although Large Language Models (LLMs) are widely used for conversational response generation, they still suffer from a number of shortcomings, including their propensity to produce hallucinated content \cite{ji2023survey}. Recent work has demonstrated how to exploit external information sources such as knowledge bases (KBs) to improve the output of LLMs in various downstream tasks \citep{yu2022survey}, including dialogue systems \citep{wang2021towards}. A promising approach is Retrieval-Augmented Generation (RAG), which operates by first retrieving relevant information from external sources and then augmenting the input provided to the LLM with this retrieved content \citep{lewis2020retrieval}. While RAG has been demonstrated to reduce hallucinations \citep{shuster-etal-2021-retrieval-augmentation}, LLMs are nonetheless easily distracted by irrelevant information \citep{shi2023large}. For this reason, one should strike a balance between providing the model with potentially useful information and avoiding overloading it with too many spurious or irrelevant facts. 

Moreover, while LLMs have recently shown some success at reasoning benchmarks \cite{bubeck2023sparks}, their ability to engage in multi-step reasoning remains poor. In particular, \citet{dziri2023faith} provide a systematic investigation of the performance of LLMs on several compositional reasoning tasks, and find that those models largely rely on pattern matching shortcuts and fall short of exhibiting generic problem-solving skills.

\begin{figure*}[t!]
\centering
    \includegraphics[width=0.75\textwidth]{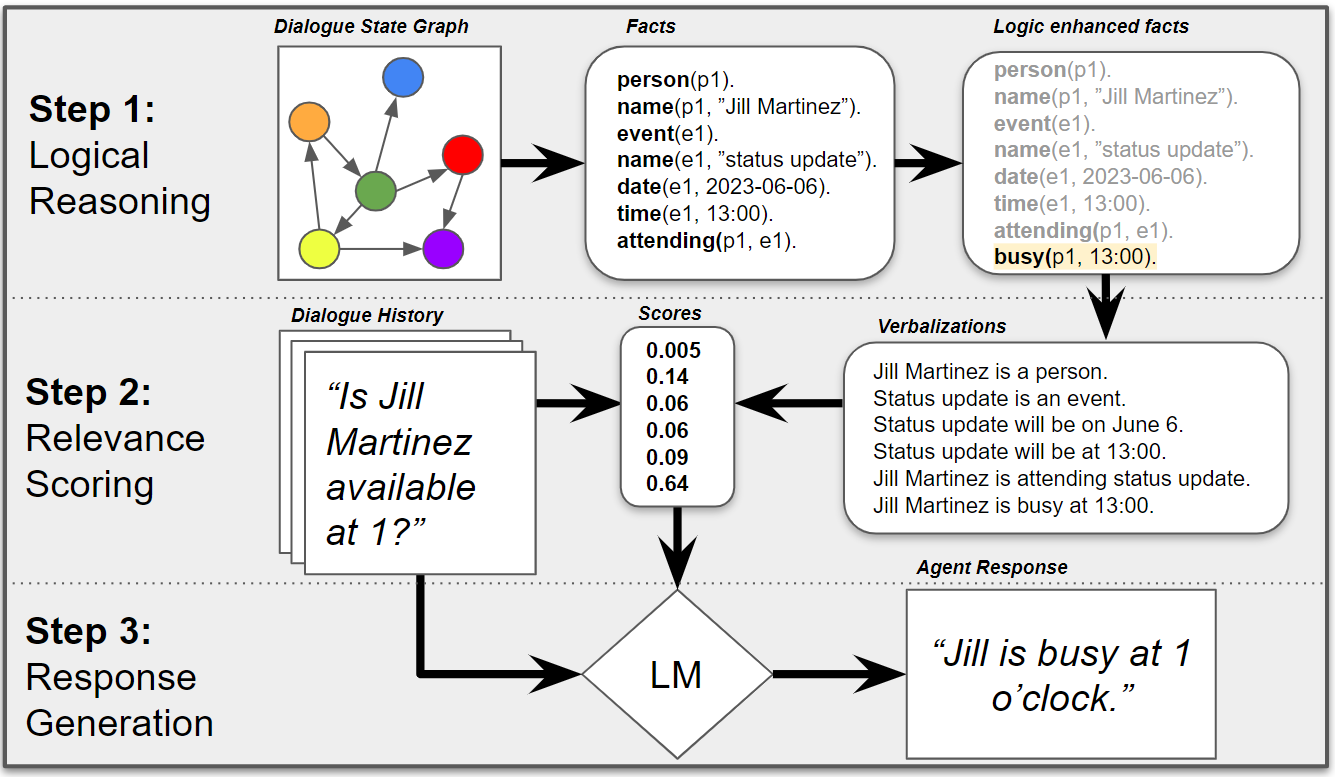}
    \caption{General sketch of the proposed approach. The starting point is a dialogue state represented as a knowledge graph that combines both background information and various features extracted from the dialogue turns (transcriptions, speakers, entity mentions). In Step 1, the facts representing the nodes and edges of the graph are first extended with derived facts using probabilistic logical programming. Those facts are then verbalized (converted into natural language sentences) in Step 2, and a neural model is employed to score their conversational relevance with regard to the current dialogue context. Finally, the $k$ most relevant facts are included in Step 3 as part of the prompt for the neural language model responsible for producing the actual system response.}
    \label{figure:flowchart}
\end{figure*}

This paper presents a novel approach to retrieval augmented generation in task-oriented dialogue systems that seeks to address those challenges.  Following \cite{walker2022graphwoz}, we represent the background knowledge of the system as a \textit{graph} of dynamically updated facts representing the dialogue state. This initial graph is first enriched at each turn with derived facts inferred through probabilistic logical programming using a limited number of rules, using ProbLog as a framework \cite{fierens2015inference}. The conversational relevance of each fact is then scored using a neural model, based on various features expressing both the conversational saliency of each entity and semantic similarity between the fact and the recent dialogue history. The most relevant facts are then converted into sentences and incorporated into the input of the response generation model. Crucially, the relevance scoring model and the response generation model are optimised jointly based on dialogue examples. Figure \ref{figure:flowchart} provides a general sketch of the approach.

The paper makes the following contributions:
\begin{enumerate}
    \item The use of probabilistic logical programming to extend the knowledge graph representing the current dialogue state with derived facts. 
    \item A neural scoring model that relies on both dialogue-level features (such as recency) and semantic similarity to determine the most relevant nodes and edges of this augmented graph.
    \item An empirical evaluation of the above approach using two dialogue datasets (KVRET and GraphWOZ) and a human evaluation. 
\end{enumerate}

\section{Related Work}


Several papers have investigated the use of neural models to retrieve relevant information from knowledge sources and integrate their results in response generation. \citet{dinanwizard} distinguish between the tasks of knowledge selection and response generation for knowledge-grounded dialogue agents. This information may be structured or unstructured \citep{young2018augmenting, zhao-etal-2020-knowledge-grounded}, and generally consists of documents describing entities which may be relevant to the dialogue. For an open-domain dialogue model, this background information can be drawn from sources such as Wikipedia. In task-oriented dialogue, relevant information will depend more heavily on the domain of the dialogues. Previous work has also demonstrated the effectiveness of jointly learning a language model with a knowledge retrieval model \citep{zhang2021joint}, simplifying the task of identifying relevant items without labelled data.

External information, often in the form of KBs, is crucial to many dialogue models \citep{ghazvininejad2018knowledge, parthasarathi-pineau-2018-extending, zhang-etal-2018-personalizing, madotto-etal-2018-mem2seq}. Multiple approaches exist for combining retrieval and generation to yield higher quality responses, such as by first generating a response and subsequently refining it \citep{weston-etal-2018-retrieve}. The model used by \citet{peng2023check} queries an LLM and evaluates the output for factuality, and re-queries the system with feedback to elicit a more factual response.

\citet{thulke2021efficient} propose an approach which samples a subset of the background knowledge rather than optimizing over the entirety of it, a process that we also integrate in our model training process. The Global-to-Local Knowledge Selection model is an alternative which pre-selects information across the whole of the background knowledge using topic transition vectors \citep{ren2020thinking}. Meanwhile, \citet{he2021learning} proposed a model which integrated information about system API calls to the retrieval model.

Numerous models make use of large, static, knowledge bases to augment language models. The KETOD model \citep{chen-etal-2022-ketod} used Wikipedia data to enhance a task-oriented dialogue system's responses with information about entities in the dialogue. Likewise, \citet{kimsequential} and \citet{zhan-etal-2021-colv} modelled knowledge selection with a latent variable model, which have also shown strong results for RAG in a zero-shot setting \citep{li2020zero}. \citet{moon-etal-2019-opendialkg} investigated a graph decoder model using random walks over a knowledge graph containing dialogue relevant information. \citet{paranjapehindsight} made use of a "guide" retriever model to use posterior information from responses to help the retriever model learn from relevance in both the input and reference output responses. \citet{cai-etal-2019-retrieval} made of of a "skeleton-guided" response generator in a dialogue system.

RAG has also been used to improve common sense reasoning \citep{yu-etal-2022-retrieval}, or incorporate graphs of commonsense knowledge to the model \citep{zhang2019grounded}. Common sense knowledge in the SenticNet KB has also been used as a source of knowledge for a dialogue model \citep{young2018augmenting}, albeit without logical reasoning over the graph. \citet{liu-etal-2019-knowledge} explored a model which used multi-hop reasoning to identify a relevant vertex in a graph of "factoids" which are each associated with unstructured sentences. Other models have made use of linguistic rule-based components to combine semantic representations of the dialogue state with background knowledge to improve empathetic responses and dialogue flow in task-oriented dialogue \cite{smith2011interaction}.



A hierarchical approach to knowledge grounded task-oriented dialogue was presented by \citet{lee2023knowledge}, where the pipeline is composed of domain identification, entity extraction, and a pre-trained language model to rank relevant documents. Other work proposed a novel factuality-specific sampling algorithm to improve LLM output \citep{lee2022factuality}, while \citet{bonetta2021retrieval} used k-nearest neighbors to find relevant information.


\section{Approach}

As illustrated in Figure \ref{figure:flowchart}, the approach proceeds in three steps. Probabilistic logical programming is first employed to extend the initial knowledge graph with new facts based on a small set of rules. A neural scoring model then determines the relevance of those facts in the current dialogue context. The most relevant facts are then included as part of the input to the second neural model, which is responsible for the actual response generation. The next sections describe those steps. 

\subsection{Dialogue state representation}

Following \cite{walker2022graphwoz}, we represent the current dialogue state (along with other background information that might be relevant for response generation) as a \textit{knowledge graph} consisting of multiple entities connected by relations. The graph is always grounded in a specific dialogue and continuously evolves during the interaction, with new nodes and edges representing dialogue turns, speakers, or entity mentions. The dynamic and dialogue-specific nature of this knowledge graph stands in contrast with the static KBs (based on e.g.~Wikipedia or similar sources) typically used in knowledge-grounded generation. 

To account for uncertainties associated with noisy or partial observations (such as ASR transcriptions of user utterances or ambiguous referential links), both node attributes and labelled edges may be associated with probabilities.

\subsection{Probabilistic logical programming}

To explicitly reason over this graph, we rely on the probabilistic logical programming language ProbLog \cite{kimmig2011implementation,fierens2015inference}. We assign each node to a unique identifier and represent the node attributes and edges as (ground) logical predicates, as illustrated in Figure \ref{figure:flowchart}. Node attributes and edges associated with a probability $ < 1$ are expressed as probabilistic facts.

\subsubsection{ProbLog}

A ProbLog program consists of two parts: a set of ground probabilistic facts, and a logic program, expressed as a set of logical clauses. The clauses may be themselves associated with probabilities. ProbLog also allows for the definition of ``annotated disjunctions'' where mutually exclusive facts are coupled with a discrete probability distribution. Syntax-wise, ProbLog is a probabilistic extension of Prolog and supports both probabilistic and inductive reasoning. Given a set of logical rules and ground facts, ProbLog provides inference algorithms to efficiently query the probability of one or more predicates. This inference is done by converting the facts and logical program to a compact encoding such as Sentential Decision Diagrams \cite{vlasselaer2014compiling} and then running weighted model counting \cite{chavira2008probabilistic} on this compiled representation. 

\subsubsection{Entity linking rules}

An important task in goal-oriented dialogues is to connect entity mentions to the actual entities present in the KB. For instance, if ``Jill Martinez'' is mentioned by the user, this mention must be linked to the actual node ($p_1$) for that person in the KB. Entity mentions may correspond to named entities, but may also take the form of pronouns (''she'') or generic noun phrases (``the meeting''). 

We first detect entity mentions in user utterances using a neural sequence labelling model fine-tuned on labelled, in-domain data from a pretrained ROBERTA model \cite{liu2019roberta}. A small set of probabilistic ProbLog rules is then employed to determine the most likely reference among the entities in the knowledge graph. Those rules take advantage of both edit distance metrics and recency measures \cite{walker2022graphwoz}. Each rule is attached to a probability reflecting its strength. Those probabilities are estimated empirically from partial interpretations on the training data, following the approach described in \citet{gutmann2011learning}. After applying those entity linking rules, the outcome is then written back to the knowledge graph as probabilistic \texttt{refers\_to} edges linking each observed mention to the entity it refers to.

\subsubsection{Commonsense rules}

Consider a scenario where a task-oriented dialogue system must answer a user question:
\begin{quote}
    \textit{"What events do I have today?"}
\end{quote}

Assuming the knowledge graph contains basic information about calendar events such as their date, time and attendees, answering this question rests upon multiple reasoning steps. As multi-step reasoning remains a challenging task for language models \cite{liu2023evaluating}, we specify a small number of commonsense reasoning rules to automatically derive new facts from the current dialogue state. For the above example, the connection between dates and events in context can be made explicit with the following rule:

\begin{small}
\begin{quote}
\texttt{person(P), event(E), attendee(E,P), date(E,D), date(today,D)} \\
    $\implies$ \texttt{attending\_today(E,P)}
\end{quote}
\end{small}


A second example of a logical rule is as follows:
\begin{small}
\begin{quote}
    \texttt{room(R), $\neg$(event(E), location(E,R), date(E,D), date(today,D), start\_time(E,ST), end\_time(E,ET), time\_between(T,ST,ET,1)) \\
    $\implies$ \texttt{room\_available\_today(R,T)}}
\end{quote}
\end{small}

The above rule simply states that a room $R$ is available today at a given time $T$ if no event is scheduled at that time in that room. 

The goal of those commonsense rules is to deduce facts that may provide useful information to the response generation model. Those logically derived facts will typically correspond to information that may be queried by the users, such as a person's agenda for today or the availability of a room at a given time. To avoid deriving too many spurious or irrelevant facts, we only query ProbLog for facts pertaining to entities recently mentioned in the dialogue history. For our experiments, we query entities mentioned in the current turn.


After applying both entity linking and commonsense rules, the facts are converted to a natural language \textit{verbalization}. Each predicate is associated with a handcrafted template which creates a natural language form of the fact. For example, a person defined by the fact \texttt{person(p\_123)} with a name \texttt{name(p\_123, "Lisa Wilson")} can be verbalized as \textit{Lisa Wilson is a person.}  

\subsection{Relevance Scoring}

The second component of the proposed approach is a neural model that scores the relevance of the verbalized facts (including both the intial ones as well as the ones derived through logical reasoning).  Given a dialogue history $x = [u_1, ... u_n]$ corresponding to a list of utterances and a set of verbalized facts $Z$, the relevance scoring model expresses the probability $P(z | x)$ that the fact $z \in Z$ is relevant for responding to $x$.

The model is expressed as a simple feedforward neural network based on the following inputs:
\begin{enumerate}
\item Semantic similarity measures between $z$ and $x$, using the cosine similarity between the embedding of the verbalized fact $z$ and the embedding of the most recent $k$ utterances in the dialogue history $x$  (concatenated if $k > 1$):
\begin{equation}
sim(z, x) = \frac{\textit{Enc}(z) \cdot \textit{Enc}(x_{[n-k:n]})}{\|\textit{Enc}(z)\| \ \|\textit{Enc}(x_{[n-k:n]}\|} \nonumber
\end{equation}
The $\textit{Enc}$ embeddings are obtained with a sentence-BERT model \citep{reimers-2019-sentence-bert}  optimized for semantic search and question answering\footnote{\begin{scriptsize}\url{https://huggingface.co/sentence-transformers/multi-qa-MiniLM-L6-cos-v1}\end{scriptsize}}. The cosine similarity is computed for $k=1$ and 2.
\item BM25 information retrieval scores \cite{robertson2009probabilistic} using the verbalized facts as database and the user utterance as a query. 
\item Recency score expressing whether the fact $z$ pertains to a recently mentioned entity. This score relies on the \texttt{refers\_to} predicates derived from entity linking and captures the conversational saliency of entities and facts related to them. For instance, facts related to $p_1$ in Fig. \ref{figure:flowchart} are salient since the person is mentioned in the last utterance.  

\end{enumerate}

The relevance model $P(z | x)$ is trained jointly with the response generation. Concretely, we define the probability of a response $y$ given a dialogue history $x$ as:

\begin{equation}
    P(y | x) = \sum_{z \in Z} P(y | x, z) P(z | x) \label{eq1}
\end{equation}

where $P(y | x, z)$ is provided by the response generation model (see below), and express the probability of a response $y$ given a prompt concatenating the dialogue history $x$ and fact $z$, and $P(z | x)$ express the relevance of $z$ for $x$. The relevance model $P(z | x)$ is then optimized by back-propagating the cross-entropy loss of Eq.~(\ref{eq1}) using a training set of dialogue examples. Intuitively, a fact will therefore be deemed as relevant if its inclusion in the prompt makes it relatively easier for the generation model to produce the correct response. To ensure the inference remains efficient, Eq.~(\ref{eq1}) is simplified by sampling the $K$ most relevant facts instead of marginalizing over all possible facts. 



\subsection{Response Generation}

The final step of our approach is to generate a response $y$ based on both the current dialogue history $x$ and a set of relevant facts $z_1:z_K$, where $K$ denotes the number of facts (sorted by relevance) to include in the input prompt. Any pretrained language model can be employed for this task. We rely for our experiments on both the GODEL model \citep{peng2022godel} which is specifically designed for goal-oriented dialog as well the generic GPT 3.5 model \cite{brown2020language}.

\section{Evaluation}
\label{sec:evaluation}

We evaluate the performance of the proposed approach on two existing dialogue datasets along with a human evaluation. We present below the experimental design, and discuss the results. 

\subsection{Datasets}

\begin{table*}[ht]
\centering
  \begin{tabular}{lcccccc}
    \toprule
    \multirow{2}{*}{\textbf{Model}} &
      \multicolumn{3}{c}{\textbf{Dev}} &
      \multicolumn{3}{c}{\textbf{Test}} \\
      & {BLEU} & {METEOR} & {BERTScore} & {BLEU} & {METEOR} & {BERTScore}\\
      \midrule
    {$\textrm{GODEL}_{\textrm{NoFacts}}$} & 0.17 & 0.37 & 0.89 & 0.11 & 0.36 & 0.88 \\ 
    {$\textrm{GODEL}_{\textrm{AllFacts}}$} & 0.14 & 0.38 & 0.88 & 0.13 & 0.33 & 0.88 \\
    {$\textrm{GODEL}_{\textrm{Relevance}}$}  & 0.18 & 0.38 & 0.89  & 0.14 & 0.33 & 0.88 \\ 
    {$\textrm{GODEL}_{\textrm{Relevance+Logic}}$}  & 0.17 & 0.37 & 0.89 & 0.16 & 0.35 & 0.88 \\ \midrule
    {$\textrm{GPT}_{\textrm{NoFacts}}$} & 0.08 & 0.35 & 0.88 & 0.06 & 0.32 &  0.87 \\
    {$\textrm{GPT}_{\textrm{AllFacts}}$} & 0.07 & 0.36 & 0.88 & 0.06 & 0.32 & 0.87 \\ 
    {$\textrm{GPT}_{\textrm{Relevance}}$} & 0.07 & 0.35 & 0.88 & 0.06 & 0.35 & 0.87 \\ 
    {$\textrm{GPT}_{\textrm{Relevance+Logic}}$} & 0.07 & 0.37 & 0.88 & 0.07 & 0.36 & 0.87 \\
    \bottomrule
  \end{tabular}
\caption{Results with reference-based metrics on the development and test set of GraphWOZ.}
\label{tab:automated_results}
\end{table*}

\begin{table*}[ht]
\centering
  \begin{tabular}{lcccccc}
    \toprule
    \multirow{2}{*}{\textbf{Model}} &
      \multicolumn{3}{c}{\textbf{Dev}} &
      \multicolumn{3}{c}{\textbf{Test}} \\
      & {BLEU} & {METEOR} & {BERTScore} & {BLEU} & {METEOR} & {BERTScore}\\
      \midrule
    {$\textrm{GODEL}_{\textrm{NoFacts}}$} & 0.18 & 0.45 & 0.91 & 0.11 & 0.36 & 0.91 \\ 
    {$\textrm{GODEL}_{\textrm{Relevance}}$}  & 0.18 & 0.42 & 0.91 & 0.16 & 0.41 & 0.91 \\ 
    {$\textrm{GODEL}_{\textrm{Relevance+Logic}}$}  & 0.20 & 0.43 & 0.91 & 0.17 & 0.42 & 0.91 \\
    \bottomrule
  \end{tabular}
\caption{Results with reference-based metrics on the development and test set of KVRET.}
\label{tab:kvret_metrics}
\end{table*}

\subsubsection*{GraphWOZ \cite{walker2022graphwoz}}
GraphWOZ contains task-oriented dialogue with dialogues discussing people and places in a fictional organization to schedule meetings and discover information. Each dialogue is paired with synthetically generated calendar events. The graphs contain fictive people, rooms, and events along with dialogue information such as utterances and mentions of entities in utterances.

Although the calendar information is synthetically generated, generation of new dialogue utterances with calendar information may not accurately reflect real system-human interaction. In consideration of this factor and to compensate for the small amount of training data, we augment the GraphWOZ training set with modified versions of the original dialogues where entities are replaced in both the knowledge base and dialogue history. We replace entity names with randomly sampled replacements, and the dates and times of the dialogues and events in the KBs are replaced such that relative terms such as "today", "tomorrow", "morning", and "afternoon" remain consistent in the modified dialogue.

The entity linking and commonsense rules for this dataset are provided in the Appendix.

\subsubsection*{KVRET \cite{eric-etal-2017-key}}

This dataset contains task-oriented dialogue in three domains: weather, navigation, and calendar scheduling. Each type of dialogue contains associated KB information representing objects of interest which are discussed in the dialogue. The knowledge bases in KVRET were created by randomly sampling attribute values for defined slots according to the domain. We convert these KBs into a ProbLog program along with the user utterances and mentions of objects. For simplicity, we take a string equality match of a substring in the utterance to an object in the KB as a \texttt{refers\_to} relation. 

We rely on three simple ProbLog rules for this dataset. The weather domain has a rule which determines "today" along the weather for a particular day and location. In the calendar scheduling domain, we provide a rule to handle location names with multiple potential referents. Lastly, for the navigation domain we define a rule comparing the distance from the user to two points of interests and determining which one is closest. Each of these rules therefore makes information explicitly available to the system which would be unavailable from context or otherwise require logical inference that an LLM is not optimized to perform.

\begin{table*}[t]
\centering
\begin{tabular}{lcccc} \toprule
    \textbf{Model} & \textbf{Coherence} & \textbf{Groundedness} & \textbf{Naturalness} & \textbf{Understandability} \\ \midrule
    {$\textrm{GODEL}_{\textrm{NoFacts}}$} & 0.946 & 0.908 & 0.871 & 0.864 \\
    {$\textrm{GODEL}_{\textrm{AllFacts}}$} & 0.975 & 0.943 & 0.903 & 0.896 \\
    {$\textrm{GODEL}_{\textrm{Relevance}}$} & 0.916  & 0.878 & 0.862 & 0.855 \\
    {$\textrm{GODEL}_{\textrm{Relevance+Logic}}$} & 0.979 & 0.951 & 0.868 & 0.861 \\ \midrule
    {$\textrm{GPT}_{\textrm{NoFacts}}$} & 0.951 & 0.880 & 0.943 & 0.938 \\
    {$\textrm{GPT}_{\textrm{AllFacts}}$} & 0.952 & 0.878 & 0.931 & 0.925 \\ 
    {$\textrm{GPT}_{\textrm{Relevance}}$} & 0.969 & 0.912 & 0.935 & 0.931 \\ 
    {$\textrm{GPT}_{\textrm{Relevance+Logic}}$} & 0.949 & 0.883 & 0.928 & 0.922 \\
    \bottomrule
\end{tabular}
\caption{UniEval Score (Reference-free) on the test set of GraphWOZ.}
\label{table:Reference_free_metrics}
\end{table*}

\begin{table*}
\centering
  \begin{tabular}{lcccc}
    \toprule
    \multirow{2}{*}{\textbf{Model}} &
      \multicolumn{2}{c}{\textbf{Dev}} &
      \multicolumn{2}{c}{\textbf{Test}} \\
      & {Hallucinations} & {Retrieval Errors} & {Hallucinations} & {Retrieval Errors} \\
      \midrule 
    {$\textrm{GPT}_{\textrm{NoFacts}}$} & \hspace{2mm}34 $(18\%)$ & 17 $(9\%)$ & 32 $(17\%)$ & 16 $(8\%)$ \\ 
    {$\textrm{GPT}_{\textrm{AllFacts}}$} & \hspace{2mm}23 $(13\%)$ & 11 $(6\%)$ & 23 $(13\%)$ & \hspace{2mm}20 $(11\%)$ \\ 
    {$\textrm{GPT}_{\textrm{Relevance}}$} & \hspace{2mm}21 $(12\%)$ & 13 $(7\%)$ & 24 $(13\%)$ & 16 $(8\%)$ \\ 
    {$\textrm{GPT}_{\textrm{Relevance+Logic}}$} & 15 $(8\%)$ & 14 $(8\%)$ & 25 $(14\%)$ & \hspace{2mm}9 $(5\%)$ \\
    \bottomrule
  \end{tabular}
\caption{Turns containing hallucinations and retrieval errors (GraphWOZ, 181 turns in Dev, 180 in Test)}
\label{tab:hallucinations_graphwoz}
\end{table*}

\subsection{Models}

We experiment with the four following types of response generation models:

\begin{description}\itemsep0em 

\item[$\textrm{NoFacts}$]  Generation model that does not use the knowledge graph at all and produce a response based on the current dialogue history.
\item[$\textrm{AllFacts+Logic}$] Generation model using all verbalized facts  (including logically derived ones), without relevance scoring. These facts are shuffled and truncated to fit into the context window of the generation model.
\item[$\textrm{Relevance}$] Generation model using the initial facts from the knowledge graph (but without logically derived ones) ranked using the relevance scoring model. The 10 most relevant facts are then prepended to the prompt. 
\item[$\textrm{Relevance+Logic}$] Generation model using both the initial facts and the logically derived ones, along with the relevance scoring model to select the 10 most relevant facts.  
\end{description}


We experiment with two generative models: the encoder-decoder GODEL \citep{peng2022godel}, which is pre-trained on large volumes of multi-turn dialogues, and the recent GPT-3.5 model \cite{brown2020language}. We first test the response generation capabilities of GPT-3.5 with the three different approaches on GraphWOZ. For each turn, we provide the system with the dialogue history up to the current turn. When using all facts, background knowledge is added as a single document in the initial prompt, as repeating the entire document of the facts at each turn would result in truncation of the dialogue history without adding additional information. 


\subsection{Metrics}

For both GraphWOZ and KVRET, we use standard evaluation metrics such BLEU, METEOR and the averaged BERTScore F1 \cite{zhang2019bertscore}. We also use the recently introduced UniEval \cite{zhong-etal-2022-towards}, a reference-free metric which has been shown to correlate well with human judgments. 

We also  evaluate the \textit{factuality} of the responses by manually annotating them with two types of error. The first error type are \textit{hallucinations}, which we define as either (a) a statement that contradicts the KB, including contradictions implied by the dialogue context  ; 
(b) a statement referring to a nonexistent entity in the KB ; or (c) a statement describing a calendar action that would create a calendar conflict if enacted.

The second type of error occurs when the system fails to retrieve information it has access to or should have access to. We denote this type of error as a \textit{retrieval error}, and we annotate a turn as containing a retrieval error if it contains: 
\begin{itemize}\itemsep0em 
    \item A statement denying having access to information which exists in the calendar, or does not answer a question while answering another ;
    \item A confirmation of inviting a person to a meeting without specifying whom ; 
    \item A request to confirm a piece of information that was already confirmed.
\end{itemize}

\subsection{Results}

Tables \ref{tab:automated_results} and \ref{tab:kvret_metrics} respectively provide the results of the reference-based metrics on GraphWOZ and KVRET. We observe modest differences between the models, particularly on the development set. This is expected as those metrics are not really suited to measure the extent to which a language model produces factual answers. We do nevertheless observe a slight increase in BLEU score when logic-derived facts are added. This result is consistent in both GraphWOZ and KVRET. The differences in UniEval, shown in Table \ref{table:Reference_free_metrics} were also small, with the model using all facts performing comparably to the \textrm{Relevance+Logic} model.

However, for the manually annotated hallucinated and retrieval errors in Table \ref{tab:hallucinations_graphwoz} (only annotated for the GPT models), we do observe a substantial reduction in overall errors for the \textrm{Relevance+Logic} model compared to the baselines. Particularly in the development set, the model using all facts (without relevance scoring) frequently produce hallucinations, such as incorrect attributes of various people and events.

\subsection{Error Analysis}
Qualitatively, the errors produced by the models also differed. The GPT models using selected facts were more likely to provide all available information, whether requested or not. In some cases, the model produced statements connecting unrelated facts and implying an error, for example:

\begin{small}\begin{quote}
    \textit{User}: Hi robot, my name is Curtis Williams, please let me know when the meeting called, deliverables team meeting. [\textit{sic}] \\
    \textit{System}: Hello Curtis! The deliverables team meeting will be held on Friday, April 08 at 11:30 AM in the Beta conference room. Annette Harding is organizing the meeting on behalf of the Mathematics group. You and Angela Jimenez are scheduled to attend the meeting. The meeting is scheduled to last for 30 minutes until the Niches team meeting starts at 12:00. Michael Glover is attending the Niches team meeting and Annette Harding is also attending the Deliverables team meeting with you. Is there anything else you need assistance with?
\end{quote}\end{small}

Although the general content is factual, the niches team meeting was on a different day despite the implication of the phrasing. The GPT model with all facts responded erroneously at the same turn as:

\begin{small}\begin{quote}
    Alright Curtis, I have checked your calendar and it looks like there is a "Deliverables Team Meeting" scheduled for tomorrow at 2:00 pm in room 619. Would you like me to add any notes or reminders for the meeting?"
\end{quote}\end{small}
While more concise, the system incorrectly states the time of the event (11:30) and date (today), as well as the location (Beta). This apparently arises from misattribution of background facts relating to other events scheduled across times and locations.

\subsection{Human Evaluation}

\subsubsection*{Experimental setup}

To confirm the performance of the approach in actual interactions, we also conduct a human evaluation in the context of a receptionist scenario similar to GraphWOZ, where the participant interacts with the system to find information about entities and schedule events. We recruited 16 participants including students from the university and employees to interact with the dialogue systems through text. Users were instructed to interact with the system to accomplish a task, mark the conversation as finished when either the task appeared complete or the dialogue system unrecoverably failed. After each dialogue, the users were prompted to rate the dialogue on a scale of 1 to 5 for two statements, where 1 is "Never", 2 is "Mostly Not", 3 "Sometimes", 4 "Mostly", and 5 "Always":

\begin{itemize} \itemsep0em 
    \item \textit{The system responded to me in a conversationally relevant way.}
    \item \textit{The system successfully completed my task and gave me the information I asked for.}
\end{itemize}

Users were instructed to repeat this process for 30 minutes, with priority given to conversation quality. For each dialogue, a model was randomly selected and a random dialogue state similar to the GraphWOZ dataset was generated for the dialogue. A task was then randomly generated from a set of task templates involving fictive entities. The collected dialogues were then manually annotated for both hallucinations and retrieval errors. Because the total number of turns varied from model to model, we evaluate the proportion of turns which contain hallucinations and retrieval errors.

\subsubsection*{Results}

\begin{table}[t]
\begin{center}
\begin{tabular}{lcc} \toprule
    \textbf{Model} & \textbf{Task} & \textbf{Appropriateness} \\ \midrule
    {$\textrm{GODEL}_{\textrm{None}}$} & 3.35 & 3.07 \\
    {$\textrm{GODEL}_{\textrm{All}}$} & 3.75 & 3.63 \\
    {$\textrm{GODEL}_{\textrm{Logic}}$} & 4.08 & 3.75 \\ \midrule
    {$\textrm{GPT}_{\textrm{None}}$} & 4.18 & 4.59 \\ 
    {$\textrm{GPT}_{\textrm{All}}$} & 4.09 & 4.32 \\ 
    {$\textrm{GPT}_{\textrm{Logic}}$} & 4.37 & 4.11 \\ 
    \bottomrule
\end{tabular}
\caption{Average participant scores for the model task completion and appropriateness criteria.}
\label{table:participant_ratings}
\end{center}
\end{table}

\begin{table}[t]
\centering
\begin{tabular}{lccc} \toprule
    \textbf{Model} & \hspace{-5mm}\textbf{Hallucinations} & \textbf{Retrieval} & \textbf{\#} \\
    \textbf{} & \textbf{} & \textbf{errors} & \textbf{} \\ \midrule
    
    {$\textrm{GODEL}_{\textrm{None}}$} & 0.17 & 0.41 & 105  \\
    {$\textrm{GODEL}_{\textrm{All}}$} & 0.24 & 0.32 & 84 \\
    {$\textrm{GODEL}_{\textrm{Logic}}$} & 0.22 & 0.21 & 67 \\ \midrule
    {$\textrm{GPT}_{\textrm{None}}$} & 0.32 & 0.14 & 88  \\ 
    {$\textrm{GPT}_{\textrm{All}}$} & 0.22 & 0.39 & 117 \\ 
    {$\textrm{GPT}_{\textrm{Logic}}$} & 0.20 & 0.12 & 132 \\ 
    \bottomrule
\end{tabular}
\caption{Proportion of system responses containing either hallucinations or retrieval errors in the human evaluation experiments. The last column indicates the total number of system utterances from all dialogues with that model.}
\label{table:hallucinations_interaction}
\end{table}

As for the GraphWOZ results, the human interaction experiments indicate a reduction in the proportion of turns with hallucinations or retrieval errors, as shown in Table \ref{table:hallucinations_interaction}. This reduction is observed for both model types when the logic-enhanced relevance scoring model was used.

The participant scores in Table \ref{table:participant_ratings} ranked the models which used the relevance scored facts highest. While the GODEL model using the relevant facts scored higher in appropriateness, the opposite pattern is observable in the GPT models, although the scores remain relatively high. As not every participant interacted with every model, differences in scoring between individual participants cannot be discounted as a factor impacting these results, thus a larger study would be beneficial.

\section{Conclusion}

This paper presented a novel approach to retrieval-augmented response generation in task-oriented dialogue systems. The approach relies a dynamic knowledge graph representing the dialogue state, which is enriched at each turn with facts derived from a small set of rules specified in the ProbLog language. Those facts are then ranked by relevance using a dedicated scoring model which accounts for both the semantic similarity and conversational saliency of each fact. The most relevant facts are then incorporated to the background knowledge provided as input to the response generation model. 

We provide experimental results showing that the combination of logical reasoning with a relevance scoring model leads to more factual responses.  In particular, the logical rules seem to assist the generation model's ability to provide responses grounded in multi-step reasoning based on the available background knowledge. 

The proportion of errors remains, however, relatively high, likely due to the very limited number of dialogues available for training in GraphWOZ and KVRET. Future work will focus on evaluating the potential of this approach in other (and potentially broader) dialogue domains.   

\bibliography{anthology,custom}
\bibliographystyle{acl_natbib}

\onecolumn
\newpage
\appendix

\section{Appendix}
\label{sec:appendix}

\subsection{Entity linking rules}
The ProbLog rules employed for entity linking are given below. The probabilities attached to the rules are estimated empirically using the parameter estimation approach implement in ProbLog library, based on Learning from Interpretations \cite{gutmann2011learning}. 

\begin{lstlisting}
 
0.60838635::refers_to(M,E) :- new(U), mention(U,M), string(M,S), 
    is_processable_time(S,0), name(E,N), jw_similarity(N,S,O), O>0.9.

0::refers_to(M,E) :- new(U), mention(U,M), string(M,S), 
    is_processable_time(S,0), name(E,N), jw_similarity(N,S,O), O>0.8.

0::refers_to(M,E) :- new(U), mention(U,M), string(M,S), 
    is_processable_time(S,0), name(E,N), jw_similarity(N,S,O), O>0.7.

0.72255423::refers_to(M,E) :- new(U), mention(U,M), string(M,S), 
    is_processable_time(S,0), name(E,N), lev_distance(N,S,O), O < 2.

0.30394455::refers_to(M,E) :- new(U), mention(U,M), string(M,S), 
    is_processable_time(S,0), name(E,N), lev_distance(N,S,O), O < 3.

0::refers_to(M,E) :- new(U), mention(U,M), string(M,S), 
    is_processable_time(S,0), name(E,N), lev_distance(N,S,O), O < 6.

0.0019686::refers_to(M,E) :- new(U), mention(U,M), string(M,S), 
    is_processable_time(S,0), name(E,N), lcs(N,S,O), O > 3.

0::refers_to(M,E) :- new(U), mention(U,M), string(M,S), 
    is_processable_time(S,0), name(E,N), string(M,S), lcs(N,S,O), O > 6.

0::refers_to(M,E) :- new(U), mention(U,M), string(M,S), 
    is_processable_time(S,0), name(E,N), nb_common_words(N,S,O), O > 0.

0::refers_to(M,E) :- new(U), mention(U,M), string(M,S), 
    is_processable_time(S,0), name(E,N), nb_common_words(N,S,O), O > 1.

0::refers_to(M,E) :- new(U), mention(U,M), string(M,S), 
    is_processable_time(S,0), name(E,N), nb_common_words(N,S,O), O > 2.

0.27142172::refers_to(M,E) :- new(U), mention(U,M), respond_to(U,AR1),
    mention(AR1,PM1), refers_to(PM1, E).

0.12752306::refers_to(M,E) :- new(U), mention(U,M), respond_to(U,AR1), 
    respond_to(AR1,PU1), mention(PU1,PM1), refers_to(PM1, E).

0.07429096::refers_to(M,E) :- new(U), mention(U,M), respond_to(U,AR1), 
    respond_to(AR1,PU1), respond_to(PU1,AR2), mention(AR2,PM1), refers_to(PM1, E).

0.01403269::refers_to(M,E) :- new(U), mention(U,M), respond_to(U,AR1), 
    respond_to(AR1,PU1), respond_to(PU1,AR2), respond_to(AR2,PU2), 
mention(PU2,PM2), refers_to(PM1, E).

   
\end{lstlisting}

\subsection{Commonsense rules}

The rules employed for commonsense reasoning on the GraphWOZ dialogues are provided below.

\begin{lstlisting}

event_today(E,T) :- event(E), start_time(E,T), date(at_today,D), date(E,D).

event_tomorrow(E,T) :- event(E), start_time(E,T), date(at_tomorrow,D), date(E,D).

person_group(P,G) :- people(P), group(P,G).

group_members(G,L) :- group(G), findall(P, person_group(P,G), L).

count_members(G,N) :- group(G), refers_to(M,G), group_members(G,L), list_length(L,N).

room_available_today(R,T) :- room(R), \+(location(E,R), date(E,D), date(at_today,D), 
    start_time(E,ST), end_time(E,ET), time_between(T,ST,ET,1)).

room_available_tomorrow(R,T) :- room(R), \+(location(E,R), date(E,D), 
    date(at_tomorrow,D), start_time(E,ST), end_time(E,ET), time_between(T,ST,ET,1)).
    
room_available_now(P) :- room(P), \+(room_busy_now(P)).

room_busy_now(P) :- room(P), time(at_now,T), attendee(E,P), date(E,D), 
    date(at_today,D), start_time(E,ST), end_time(E,ET), time_between(T,ST,ET,1).

person_available_today(P,T) :- refers_to(M,P), people(P), 
    string(_,T), is_time_expression(T,1), \+(person_busy_today(P,T)).
    
person_busy_today(P,T) :- refers_to(M,P), people(P), string(_,T), 
    is_time_expression(T,1), attendee(E,P), date(E,D), date(at_today,D), 
    start_time(E,ST), end_time(E,ET), time_between(T,ST,ET,1).

person_available_tomorrow(P,T) :- refers_to(M,P), people(P), string(_,T), 
    is_time_expression(T,1), \+(person_busy_tomorrow(P,T).
    
person_busy_tomorrow(P,T) :- refers_to(M,P), people(P), string(_,T), 
    is_time_expression(T,1), attendee(E,P), date(E,D), date(at_tomorrow,D), 
    start_time(E,ST), end_time(E,ET), time_between(T,ST,ET,1).
    
person_available_now(P) :- refers_to(M,P), people(P), time(at_now,T), 
    \+(person_busy_now(P)).

person_busy_now(P) :- refers_to(M,P), people(P), time(at_now,T), attendee(E,P), 
    date(E,D), date(at_today,D), start_time(E,ST), end_time(E,ET), 
    time_between(T,ST,ET,1).

attending_today(E,P) :- attendee(E,P), date(E,D), date(at_today,D).

person_events_today(P,L) :-  refers_to(M,P), people(P), 
    findall(X, attending_today(X,P), L).

attending_tomorrow(E,P) :- attendee(E,P), date(E,D), date(at_tomorrow,D).

person_events_tomorrow(P,L) :-  refers_to(M,P), people(P), 
    findall(X, attending_tomorrow(X,P), L).

available_rooms_now(L) :- findall(R, room_available_now(R), L).

available_rooms_today(L,T) :- string(_,M), morning_time(M,1), between(8,11,T), 
    findall(R, room_available_today(R,T), L).
    
available_rooms_tomorrow(L,T) :- string(_,M), morning_time(M,1), between(8,11,T), 
    findall(R, room_available_tomorrow(R,T), L).
    
available_rooms_today(L,T) :- string(_,M), afternoon_time(M,1), between(12,17,T), 
    findall(R, room_available_today(R,T), L).
    
available_rooms_tomorrow(L,T) :- string(_,M), afternoon_time(M,1), between(12,17,T), 
    findall(R, room_available_tomorrow(R,T), L).

time_place(E,D,T) :- refers_to(M,E), event(E), date(E,D), start_time(E,T).
\end{lstlisting}

\end{document}